\documentclass[runningheads]{llncs}
\usepackage{graphicx}

\usepackage{tikz}
\usepackage{comment} 
\usepackage{amsmath,amssymb} 
\usepackage{color}

\usepackage{tabulary}
\usepackage{epsfig}
\usepackage{graphics}
\usepackage{mathtools}
\usepackage{xcolor}
\usepackage{multirow}
\usepackage{hhline}
\usepackage{makecell}
\usepackage{lineno}
\usepackage{booktabs}
\usepackage{hyperref} 
\usepackage{url} 
\usepackage{bigstrut}
\usepackage{array, boldline, rotating}
\usepackage[ruled,vlined,linesnumbered,boxed]{algorithm2e}

\hypersetup{
    colorlinks=true,
    linkcolor=red,
    filecolor=magenta,      
    urlcolor=cyan,
}

\makeatletter
\newcommand{\printfnsymbol}[1]{%
  \textsuperscript{\@fnsymbol{#1}}%
}
\makeatother

\begin{document}
\pagestyle{headings}
\mainmatter
\def\ECCVSubNumber{1593}  

\title{\textit{ViTAA}: Visual-Textual Attributes Alignment\\ 
in Person Search by Natural Language} 

\titlerunning{Visual-Textual Attributes Alignment in Person Search by Natural Language}
%
\author{Zhe Wang\inst{1 \thanks{Equal contribution. This work was done when Z. Wang
was a visiting scholar at Active Perception Group, Arizona State University.}} \and
Zhiyuan Fang \inst{2 \printfnsymbol{1}} \and
Jun Wang \inst{1} \and 
Yezhou Yang \inst{2}}
\authorrunning{Z. Wang et al.}

\institute{Beihang University \\
\email{\{wangzhewz,wangj203\}@buaa.edu.cn}
\and
Arizona State University \\
\email{\{zfang29,yz.yang\}@asu.edu}\\}
    
\maketitle

\begin{abstract}
Person search by natural language aims at retrieving a specific person in a large-scale image pool that matches given textual descriptions. While most of the current methods treat the task as a holistic visual and textual feature matching one, we approach it from an attribute-aligning perspective that allows grounding specific attribute phrases to the corresponding visual regions.
We achieve success as well as a performance boost by a robust feature learning that the referred identity can be accurately bundled by multiple attribute cues. 
To be concrete, our Visual-Textual Attribute Alignment model (dubbed as {ViTAA}) learns to disentangle the feature space of a person into sub-spaces corresponding to attributes using a light auxiliary attribute segmentation layer. It then aligns these visual features with the textual attributes parsed from the sentences via a novel contrastive learning loss.
We validate our ViTAA framework through extensive experiments on tasks of person search by natural language and by attribute-phrase queries, on which our system achieves state-of-the-art performances. Codes and models are available at \url{https://github.com/Jarr0d/ViTAA}.

\end{abstract}

\begin{figure}[t]
\centering
\includegraphics[width=1.0\textwidth]{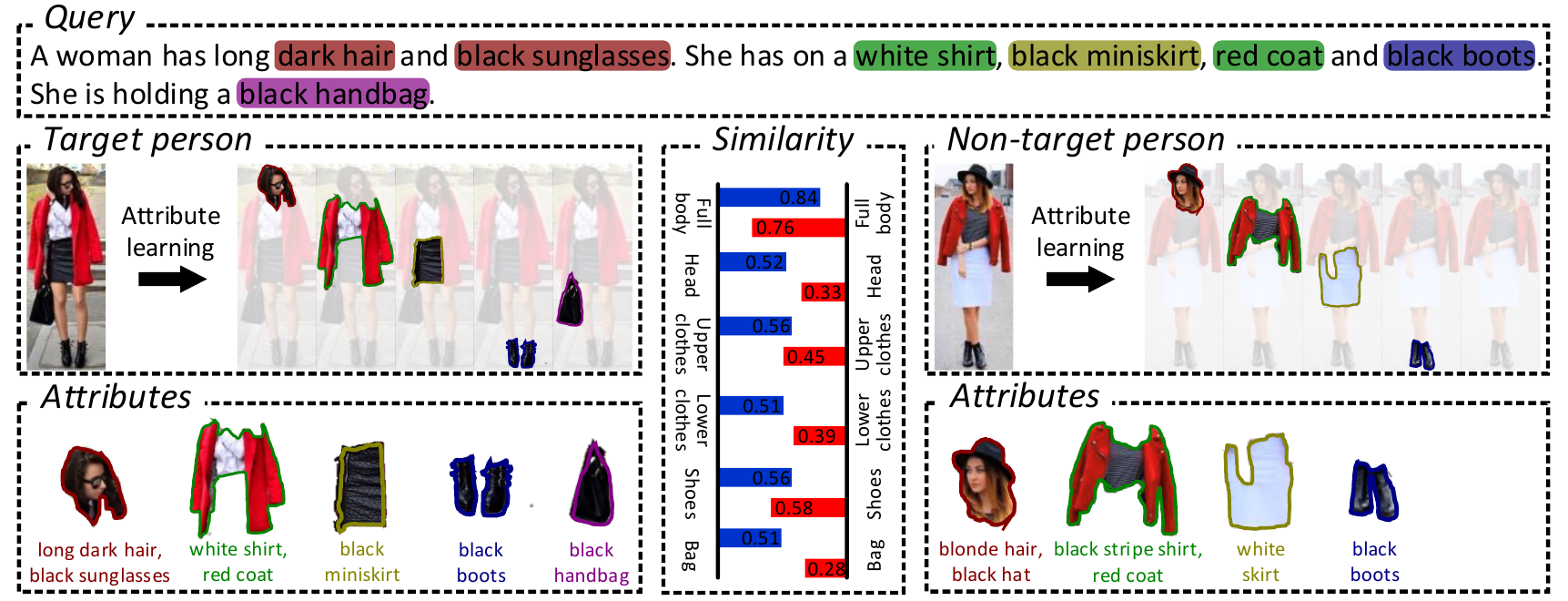}
\caption{In a case when two persons exhibit similar appearance attributes, it is hard to discriminate them merely by full-body appearance. Instead of matching the textual descriptions with the images at global-level, we decompose both image and text into attribute components and conduct a fine-grained matching strategy.}
\label{fig:modeling}
\end{figure}

\section{Introduction}
Recently, we have witnessed numerous practical breakthroughs in person modeling related tasks, \textit{e.g.}, pedestrian detection~\cite{benenson2014ten,dollar2009pedestrian,wang2020resisting}, pedestrian attribute recognition~\cite{liu2017hydraplus,sudowe2015person} and
person re-identification~\cite{Gong:2014:PR:2584512,li2014deepreid,zheng2015scalable}. Person search~\cite{li2017person,han2019re} as an aggregation of the aforementioned tasks thus gains increasing research attention. Comparing with searching by image queries, person search by natural language~\cite{li2017person,li2017identity,Dong_2019_ICCV,ijcai2018-153} makes the retrieving procedure more user-friendly with increased flexibility due to a supporting of open-form natural language queries. Meanwhile, learning robust visual-textual associations becomes increasingly critical, which calls an urgent demand for a representation learning schema that is able to fully exploit both modalities.

Relevant studies in person modeling related research points out the critical role of the discriminative representations, especially of the local fragments in both image and text. For the former, ~\cite{su2017pose,zheng2019pose} learn the pose-related features from the key points map of human, while~\cite{kalayeh2018human,liang2018look} leverage the body-part features by auxiliary segmentation-based supervision.
For the latter,~\cite{li2017person,li2017identity,zhang2018deep} decompose the complex sentences into noun phrases, and~\cite{ijcai2018-153,layne2014attributes} directly adopt the attribute-specific annotations to learn fine-grained attribute related features. 
Moving forward, attribute specific features from image and text are even requisite for person search by natural language task, and how to effectively couple them stays an open question. We seek insight from a fatally flawed case that lingers in most of the current visual-language systems in Figure~\ref{fig:modeling}, termed as ``malpositioned matching''. For example, tasks like textual grounding~\cite{rohrbach2016grounding,plummer2015flickr30k}, VQA~\cite{antol2015vqa}, and image-text retrieval~\cite{shekhar2012word,jeon2003automatic} are measuring the similarities or mutual information across modalities in a holistic fashion by answering: are the feature vectors of image and text match with each other? 
That way, when users input ``\textit{a girl in white shirt and black skirt}'' as retrieval query, the model is not able to distinguish the nuances of the two images as shown in Figure~\ref{fig:modeling}, where the false positive one actually shows ``\textit{black shirt and white skirt}''.
As both the distinct color visual cues (``\textit{white}'' and ``\textit{black}'') exist in the images, overall matching without the ability of referring them to specific appearance attributes prevents the model from discriminating them as needed. Such cases exist extensively in almost all cross-modal tasks, which pose an indispensable challenge for a system to tackle with the ability of fine-grained interplay between image and text.

Here, we put forward a novel \textbf{Vi}sual-\textbf{T}extual \textbf{A}ttributes \textbf{A}lignment model (ViTAA). For feature extraction, we fully exploit both visual and textual attribute representations. Specifically, we leverage segmentation labels to drive the attribute-aware feature learning from the input image. As shown in Figure~\ref{fig:architecture}, we design multiple local branches, each of which is responsible to predict one particular attribute visual feature. This process is guided by the supervision on segmentation annotations, so the features are intrinsically aligned through the label information. We then use a generic natural language parser to extract attribute-related phrases, which at the same time remove the complex syntax in natural language and redundant non-informative descriptions. 
Building upon this, we adopt a contrastive learning schema to learn a joint embedding space for both visual and textual attributes. Meanwhile, we also notice that there may exist common attributes across different person identities (\textit{e.g.,} two different persons may wear similar ``\textit{black shirt}''). To thoroughly exploits these cases during training, we propagate a novel sampling method to mine surrogate positive examples which largely enriches the sampling space, and also provides us with valid informative samples for the sake of overcoming convergence problem in metric learning. 

To this end, we argue and show that the benefits of the attribute alignment to person search model go well beyond the obvious. As the images used for person search tasks often contain a large variance on appearance (\textit{e.g.}, varying poses or viewpoints, with/without occlusion, and with cluttered background), the abstracted attribute-specific features could naturally help to resolve the ambiguity in feature representations. Also, searching by appearance attributes innately brings interpretability for the retrieving task and enables the attribute specific retrieval.
It is worth mentioning that, there exist few very recent efforts that attempt to utilize the local fragments in both visual and textual modalities~\cite{Fang_2019_CVPR,Wu_2019_CVPR} and hierarchically align them~\cite{Dong_2019_ICCV,chen2018improving}. The pairing schema of visual features and textual phrases in these methods are all based on the same identity, where they neglect the cues that exist across different identities. 
Comparing with them, ours is a more comprehensive modeling method that fully exploits the identical attributes from different persons thus greatly helps the alignment learning.

To validate these speculations, extensive experiments are conducted to our ViTAA model on the task of 1) person search by natural language and 2) by attribute, showing that our proposed model is capable of linking specific visual cues with specific words/phrases. More concretely, our ViTAA achieves a promising results across all these tasks. Further qualitative analysis verifies that our alignment learning successfully learns the fine-grained level correspondence across the visual and textual attributes. To summarize our contributions:
\begin{itemize}
\item We design an attribute-aware representation learning framework to extract and align both visual and textual features for the task of person search by natural language. To the best of our knowledge, this is the first to adopt both semantic segmentation as well as natural language parsing to facilitate a semantically aligned representation learning.

\item We design a novel cross-modal alignment learning schema based on contrastive learning which can adaptively highlight the informative samples during the alignment learning. Meanwhile, an unsupervised data sampling method is proposed to facilitate the construction of learning pairs by exploiting more surrogate positive samples across different person identities.

\item We evaluate the superiority of ViTAA over other state-of-the-art methods for the person search by natural language task. We also conduct qualitative analysis to demonstrate the interpretability of ViTAA.
\end{itemize}

\section{Related Work}
\textbf{Person Search.} Given the form of the querying data, current person search tasks can be categorized into two major thrusts: searching by images (termed as Person Re-Identification), and person search by textual descriptions. Typical person re-identification (re-id) methods~\cite{Gong:2014:PR:2584512,li2014deepreid,zheng2015scalable} are formulated as retrieving the candidate that shows highest correlation with the query in the image galleries. However, a clear and valid image query is not always available in the real scenario, thus largely impedes the applications of re-id tasks.
Recently, researchers alter their attention to re-id by textual descriptions: identifying the target person by using free-form natural languages~\cite{li2017person,li2017identity,chen2018improving}. Meanwhile, it also comes with great challenges as requiring the model to deal with the complex syntax from the long and free-form descriptive sentence, and the inconsistent interpretations of low-quality surveillance images. To tackle these, methods like~\cite{li2017person,li2017identity,8354312} employ attention mechanism to build relation module between visual and textual representations, while~\cite{zhang2018deep,zheng2017dual} propose cross-modal objective functions for joint embedding learning. Dense visual feature is extracted in~\cite{niu2019improving} by cropping the input image for learning a regional-level matching schema. Beyond this,~\cite{jing2018pose} introduces pose estimation information for delicate human body-part parsing.

\noindent\textbf{Attribute Representations.} Adopting appropriate feature representations is of crucial importance for learning and retrieving from both image and text. Previous efforts in person search by natural language unanimously use holistic features of the person, which omit the partial visual cues from attributes at the fine-grained level.
Multiple re-id systems have focused on the processing of body-part regions for visual feature learning, which can be summarized as: hand-craft horizontal stripes or grid~\cite{li2014deepreid,sun2018beyond,wang2018learning}, attention mechanism~\cite{si2018dual,wang2018mancs}, and auxiliary information including keypoints~\cite{xu2018attention,suh2018part}, human parsing mask~\cite{kalayeh2018human,liang2018look} and dense semantic estimation~\cite{zhang2019densely}. Among these methods, the auxiliary information usually provides more accurate partition results on localizing human parts and facilitating body-part attribute representations thanks to the multi-task training or the auxiliary networks. However, only few work~\cite{Guo_2019_ICCV} pay attention to the accessories (such as the backpack) which could be the potential contextual cues for accurate person retrieval.
As the corresponding components to specific visual cues, textual attribute phrases are usually provided as ground-truth labels or can be extracted from sentences through identifying the noun phrases with sentence parsing. Many of them use textual attributes as auxiliary label information to complement the content of image features~\cite{layne2014attributes,su2018multi,lin2019improving}. Recently, a few attempts leverage textual attribute as query for person retrieval~\cite{Dong_2019_ICCV,ijcai2018-153}.~\cite{ijcai2018-153} imposes an
attribute-guided attention mechanism to capture the holistic appearance of person.~\cite{Dong_2019_ICCV} proposes a hierarchical matching model that can jointly learn global category-level and local attribute-level embedding.

\noindent\textbf{Visual-Semantic Embedding.} Works in vision and language propagate the notion of visual semantic embedding, with a goal to learn joint feature space for both visual inputs and their correspondent textual annotations~\cite{frome2013devise,you2018end}. Such mechanism plays a core role in a series of cross-modal tasks, \textit{e.g.}, image/video captioning~\cite{karpathy2015deep,xu2015show,fang2020video2commonsense}, image retrieval through natural language~\cite{zhang2018deep,Wu_2019_CVPR,Fang_2019_CVPR}, and vision question answering~\cite{antol2015vqa}.
Conventional joint embedding learning framework adopts two-branch architecture~\cite{zhang2018deep,frome2013devise,you2018end,Fang_2019_CVPR,fang2018weakly}, where one  extracts image features and the other  encodes textual descriptions. The extracted cross-modal embedding features are learned through carefully designed objective functions.

\section{Our Approach}
Our network is composed of an image stream and a language stream (see Figure~\ref{fig:architecture}), with the intention to encode inputs from both modalities for a visual-textual embedding learning. To be specific, given a person image $\mathcal{I}$ and its textual description $\mathcal{T}$, we first use the image stream to extract a global visual representation $\boldsymbol{v}_0$, and a stack of local visual representations of $N_{att}$ attributes $\{\boldsymbol{v}_1, ... ,\boldsymbol{v}_{N_{att}}\}$, $\boldsymbol{v}_i\in \mathbb{R}^{d}$.
Similarly, we follow the language stream to extract overall textual embedding $\boldsymbol{t}_0$,
then decompose the whole sentence using standard natural language parser~\cite{klein2003fast} into a list of the attribute phrases, and encode them as $\{\boldsymbol{t}_1, ... ,\boldsymbol{t}_{N_{att}}\}$, $\boldsymbol{t}_i\in\mathbb{R}^d$. Our core contribution is the cross-modal alignment learning that matches each visual component $\boldsymbol{v}_a$ with its corresponding textual phrase $\boldsymbol{t}_a$, along with the global representation matching $\big<\boldsymbol{v}_0, \boldsymbol{t}_0\big>$ for the person search by natural language task. 

\subsection{The Image Stream}
\begin{figure*}[t!]
\centering
\includegraphics[width=1.0\linewidth]{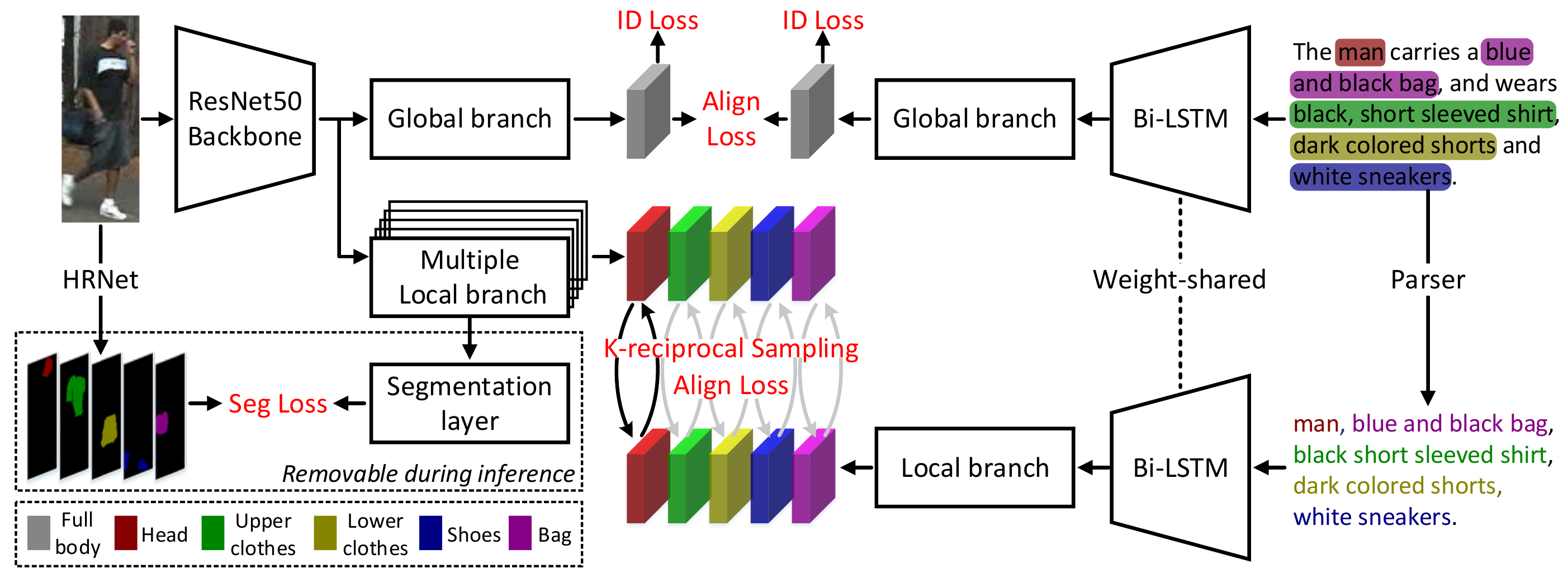}
\caption{Illustrative diagram of our ViTAA network, which includes an image stream (left) and a language stream (right). Our image stream first encodes the person image and extract both global and attribute representations. The local branch is additional supervised by an auxiliary segmentation layer where the annotations are acquired by an off-the-shell human parsing network. 
In the meanwhile, the textual description is parsed and decomposed into attribute atoms, and encoded by a weight-shared Bi-LSTM. We train our ViTAA jointly under global/attribute align loss in an end-to-end manner.}
\label{fig:architecture}
\end{figure*}

We adopt the sub-network of ResNet-50 (conv1, conv2\_x, conv3\_x, conv4\_x)~\cite{he2016deep} as the backbone to extract feature maps $\boldsymbol{F}$ from the input image. Then, we introduce a global branch $\mathcal{F}^{glb}$, and multiple local branches $\mathcal{F}^{loc}_{a}$ to generate global visual features $\boldsymbol{v}_{0} = \mathcal{F}^{glb}(\boldsymbol{F})$, 
and attribute visual features $\{\boldsymbol{v}_{1}\dots\boldsymbol{v}_{N_{att}}\}$ respectively, where $\boldsymbol{v}_{a} = \mathcal{F}^{loc}_{a}(\boldsymbol{F})$. The network architectures are shown in Table~\ref{tab:head}. 
On the top of all the local branches is an auxiliary segmentation layer supervising each local branch to generate the segmentation map of one specific attribute category (shown in Figure~\ref{fig:architecture}). 
Intuitively, we argue that the additional auxiliary task acts as a knowledge regulator that diversifies each local branch to present attribute-specific features.

\begin{figure}[t]
\centering
\includegraphics[width=1.0\textwidth]{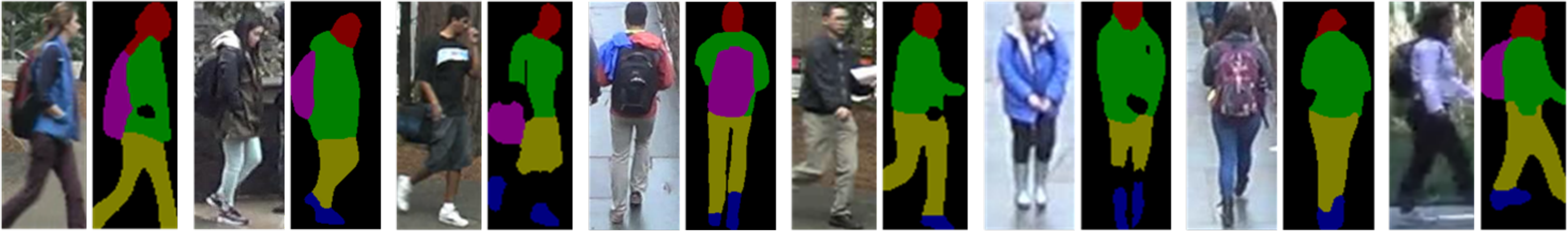}
\caption{
Attribute annotation generated by human parsing network. Torsos are labeled as background since there is no corresponding textual descriptions.
}
\label{fig:annotation}
\end{figure}

Our segmentation layer utilizes the architecture of a lightweight MaskHead~\cite{he2017mask} and can be removed during inference phase to reduce the computational cost. The remaining unsolved problem is that parsed annotations are not available in all person search datasets. 
To address that, we first train a human parsing network with HRNet~\cite{Sun_2019_CVPR} as an off-the-shelf tool, where the HRNet is jointly trained on multiple human parsing datasets: MHPv2~\cite{zhao2018understanding}, ATR~\cite{ATR}, and VIPeR~\cite{tan2019attention}.
We then use the attribute category predictions as our segmentation annotations (illustrated in Figure~\ref{fig:annotation}). With these annotations, local branches receive the supervision needed from the segmentation task to learn attribute-specific features. 
Essentially, we are distilling the attribute information from a well-trained human parsing networks to the lightweight segmentation layer through joint training\footnote[1]{More details of our human parsing network and segmentation results can be found in the experimental part and the supplementary materials.}.

\noindent\textbf{Discussion.} 
Using attribute feature has the following advantages over the global features. 1) The textual annotations in person search by natural language task describe the person mostly by their dressing/body appearances, where the attribute features perfectly fit the situation.
2) Attribute aligning avoids the ``\textit{malpositioned matching}'' cases as shown in Fig.~\ref{fig:modeling}:
using segmentation to regularize feature learning equips the model to be resilient over the diverse human poses or viewpoints, and also robust to the background noises.

\subsection{The Language Stream}
Given the raw textual description, our language stream first parses and extracts noun phrases \textit{w.r.t.} each attribute through the Stanford POS tagger~\cite{manning-EtAl:2014:P14-5}, and then feeds them into a language network to obtain the sentence-level as well as the phrase-level embeddings. We adopt a bi-directional LSTM to generate the global textual embedding $\boldsymbol{t}_0$ and the local textual embedding. 
Meanwhile, we adopt a dictionary clustering approach to categorize the novel noun phrases in the sentence to specific attribute phrases as in~\cite{Fang_2019_CVPR}.
Concretely, we manually collect a list of words per attribute category, \textit{e.g.}, ``\textit{shirt}'', ``\textit{jersey}'', ``\textit{polo}'' to represent the upper-body category, and use the average-pooled word vectors~\cite{goldberg2014word2vec} of them as the anchor embedding $\mathbf{d}_a$, and form the dictionary $\mathbf{D} = [\mathbf{d}_1, ... ,\mathbf{d}_{N_{att}}]$, where $N_{att}$ is the total number of attributes. Building upon that, we assign the noun phrase to the category that has the highest cosine similarity, and form the local textual embedding \{$\boldsymbol{t}_1\dots\boldsymbol{t}_N$\}. Different from previous works like~\cite{zhang2019densely,niu2019improving}, we include accessory as one type of attribute as well, which serves as a crucial matching clue in many cases. 


\subsection{Visual-Textual Alignment Learning}
Once we extract the global and attribute features, the key objective for the next stage is to learn a joint embedding space across the visual and the textual modalities, where the visual cues are tightly matched with the given textual description. 
Mathematically, we formulate our learning objective as a contrastive learning task that takes input as triplets, \textit{i.e.}, $\big<\boldsymbol{v}^i, \boldsymbol{t}^+, \boldsymbol{t}^-\big>$ and $\big<\boldsymbol{t}^i, \boldsymbol{v}^+, \boldsymbol{v}^-\big>$, where $i$ denotes the index of person to identify, and $+/-$ refer to the corresponding feature representations of the person $i$, and a randomly sampled irrelevant person respectively.
We note that features in the triplet can be both at the global-level and the attribute-level. In the following, we discuss the learning schema on $\big<\boldsymbol{v}^i, \boldsymbol{t}^+, \boldsymbol{t}^-\big>$ which can be extended to $\big<\boldsymbol{t}^i, \boldsymbol{v}^+, \boldsymbol{v}^-\big>$.

We adopt the cosine similarity as the scoring function between visual and textual features $S= \frac{\boldsymbol{v}^T\cdot\boldsymbol{t}}{||\boldsymbol{v}||\cdot||\boldsymbol{t}||}$. For a positive pair $\big<\boldsymbol{v}^i, \boldsymbol{t}^+\big>$, the cosine similarity $S^+$ is encouraged to be as large as possible, which we define as \emph{absolute similarity criterion}. While for a negative pair $\big<\boldsymbol{v}^i, \boldsymbol{t}^-\big>$, enforcing the cosine similarity $S^-$ to be minimal may yield an arbitrary constraint over the negative samples $\boldsymbol{t}^-$. Instead, we propose to optimize the deviation between $S^-$ and $S^+$ to be larger than a preset margin, called \emph{relative similarity criterion}. These criterion can be formulated as:
\begin{equation}
S^+\rightarrow 1\ and\ S^+ - S^- > m,
\label{eq:criterion}
\end{equation}
where $m$ is the least margin that positive and negative similarity should differ and is set to $0.2$ in practice.

In contrastive learning, the general form of the basic objective function are either hinge loss $\mathcal{L}(\boldsymbol{x})=\max\{0, 1-\boldsymbol{x}\}$ or logistic loss $\mathcal{L}(\boldsymbol{x})=\log(1+\exp{(-\boldsymbol{x})})$. One crucial drawback of hinge loss is that its derivative \textit{w.r.t.} $x$ is a constant value: $\frac{\partial \mathcal{L}}{\partial x}=-1$. Since the pair-based construction of training data leads to a polynomial growth of training pairs, inevitably we will have a certain part of the randomly sampled negative texts being less informative during training. Treating all the redundant samples equally might raise the risk of a slow convergence and/or even model degeneration for the metric learning tasks. While the derivative of logistic loss \textit{w.r.t.} $x$ is: $\frac{\partial \mathcal{L}}{\partial x}=-\frac{1}{e^x+1}$, which is related with the input value. Hence, we settle with the logistic loss as our basic objective function.

With the logistic loss, the aforementioned criterion can be further derived and rewritten as:
\begin{equation}
(S^+-\alpha) > 0,\ -( S^- - \beta) > 0,
\label{eq:derivation}
\end{equation}
where $\alpha\rightarrow 1$ denotes the lower bound for positive similarity and $\beta = (\alpha - m)$ denotes the upper bound for negative similarity. Together with logistic loss function, our final \emph{Alignment loss} can be unrolled as: 
\begin{equation}
\mathcal{L}_{align} = \frac{1}{N}\sum_{i=1}^{N}\Big\{ \log\left[1 + e^{-\tau_p(S_{i}^+ - \alpha)}\right] + \log\left[1 + e^{\tau_n(S_i^{-} - \beta)}\right] \Big\},
\label{eq:loss} 
\end{equation}
where $\tau_p$ and $\tau_n$ denote the temperature parameters that adjust the slope of gradient. The partial derivatives are calculated as:
\begin{equation}
\frac{\partial \mathcal{L}_{align}}{\partial S_i^{+}} = \frac{-\tau_p}{1 + e^{\tau_p (S_i^{+} - \alpha)}}, \frac{\partial \mathcal{L}_{align}}{\partial S_i^{-}} = \frac{\tau_n}{1+e^{\tau_n (\beta - S_i^{-})}}.
\label{eq:weight} 
\end{equation}
Thus, we show that Eq.~\ref{eq:loss} outputs continuous gradients and will assign higher weights to more  informative samples accordingly.

\noindent\textbf{K-reciprocal Sampling.}
One of the premise of visual-textual alignment is to fully exploit the informative positive and negative samples $\boldsymbol{t}^+, \boldsymbol{t}^-$ to provide valid supervisions. However, most of the current contrastive learning methods~\cite{wen2016discriminative,zhang2017range} construct the positive pairs by selecting samples belonging to the same class and simply treat the random samples from other classes as negative. This is viable when using only global information at coarse level during training, but may not be able to handle the case as illustrated in Figure~\ref{fig:modeling} where a fine-grained level comparison is needed. This practice is largely depending on the average number of samples for each attribute category to provide comprehensive positive samples. With this insight, we propose to further enlarge the searching space of positive samples from the cross-id incidents.

For instance, as in Figure~\ref{fig:modeling}, though the two ladies are with different identities, they share the extremely alike shoes which can be treated as the positive samples for learning. We term these kinds of samples with identical attributes but belong to different person identities as the ``surrogate positive samples''. Kindly including the common attribute features of the surrogate positive samples in positive pairs makes much more sense than the reverse. It is worth noting that, this is unique only to our attribute alignment learning phase because attributes can only be compared at the fine-grained level. Now the key question is, how can we dig out the surrogate positive samples since we do not have direct cross-ID attribute annotations? Inspired by the re-ranking techniques in re-id community~\cite{garcia2015person,zhong2017re}, we propose k-reciprocal sampling as an unsupervised method to generate the surrogate labels at the attribute-level. How does the proposed method sample from a batch of visual and textual features? Straightforwardly, for each attribute $a$, we can extract a batch of visual and textual features from the feature learning network and mine their corresponding surrogate positive samples using our sampling algorithm. Since we are only discussing the input form of $\big<\boldsymbol{v}^i, \boldsymbol{t}^+, \boldsymbol{t}^-\big>$, our sampling algorithm is actually mining the surrogate positive textual features for each $\boldsymbol{v}^i$. Note that, if the attribute information in either modality is missing after parsing, we can simply ignore them during sampling.

\begin{center}
\resizebox{.9\textwidth}{!}{%
\begin{algorithm}[H]
\SetAlgoLined
\KwIn{ \\
\quad $\mathcal{V}_a=\{\boldsymbol{v}^{i}_{a}\}_{i=1}^N$ is a set of visual feature for attribute $a$ \\
\quad $\mathcal{T}_a=\{\boldsymbol{t}^{i}_{a}\}_{i=1}^N$ is a set of textual feature for
attribute $a$ \\
}
\KwOut{ \\
\quad $\mathcal{P}_a$ is a set of surrogate positive sample for attribute $a$ \\
}
 \For{each $\boldsymbol{v}_{a} \in\mathcal{V}_a$}{
  find the top-$\emph{K}$ nearest neighbours of $\boldsymbol{v}_{a}$ \emph{w.r.t.} $\mathcal{T}_a$: $\mathcal{K}_{\mathcal{T}_a}$\;
  $\mathcal{S}\leftarrow\varnothing$\;
  \For{each $\boldsymbol{t}_{a} \in \mathcal{K}_{\mathcal{T}_a}$}{
  find the top-$\emph{K}$ nearest neighbours of $t_{a}$ \emph{w.r.t.} $\mathcal{V}_a$: $\mathcal{K}_{\mathcal{V}_a}$\;
   \If{$\boldsymbol{v}_{a} \in \mathcal{K}_{\mathcal{V}_a}$}{
    $\mathcal{S}=\mathcal{S}\cup \boldsymbol{t}_{a}$
   }
  }
  $\mathcal{P}_a=\mathcal{P}_a\cup \mathcal{S}$
 }
\caption{\textit{K}-\textit{reciprocal} Sampling Algorithm}
\end{algorithm}
\label{alg:sample}
}
\end{center}

\subsection{Joint Training}
The entire network is trained in an end-to-end manner. We adopt the widely-used cross-entropy loss (ID Loss) to assist the learning of the discriminative features of each instance, as well as pixel-level cross-entropy loss (Seg Loss) to classify the attribute categories in the auxiliary segmentation task. For the cross-modal alignment learning, we design the Alignment Loss on both the global-level and the attribute-level representations. The overall loss function thus emerges:
\begin{equation}
\mathcal{L} = \mathcal{L}_{id} + \mathcal{L}_{seg} + \mathcal{L}_{align}^{glo} + \mathcal{L}_{align}^{attr}.
\tag{5}
\label{eq:5} 
\end{equation}

\begin{table}[t]
\caption{Detailed architecture of our global and local branches in image stream. \emph{\#Branch.} denotes the number of sub-branches.}
\begin{center}
\setlength{\tabcolsep}{7.5pt}
\begin{tabular}{c|c|c|c}
\toprule[1pt]
\multicolumn{1}{c|}{Layer name} & \multicolumn{1}{c|}{Parameters} & \multicolumn{1}{c|}{Output size} & \#Branch \\
\hline
 \multicolumn{1}{c|}{\multirow{3}[4]{*}{$\mathcal{F}^{glb}$}} & \multicolumn{1}{c|}{\multirow{2}[2]{*}{$\begin{bmatrix}3\times3,2048\\3\times3,2048\end{bmatrix}\times2$}} & \multicolumn{1}{c|}{\multirow{2}[2]{*}{$24\times8$}} & \multicolumn{1}{c}{\multirow{3}[4]{*}{1}} \bigstrut\\
\multicolumn{1}{c|}{} & \multicolumn{1}{c|}{} & \multicolumn{1}{c|}{} & \multicolumn{1}{c}{} \\
\cline{2-3}
\multicolumn{1}{c|}{} & Average Pooling & $1\times1$ & \bigstrut \\
\hline
 \multicolumn{1}{c|}{\multirow{3}[4]{*}{$\mathcal{F}^{loc}$}} & \multicolumn{1}{c|}{\multirow{2}[2]{*}{$\begin{bmatrix}3\times3,256\\3\times3,256\end{bmatrix}\times2$}} & \multicolumn{1}{c|}{\multirow{2}[2]{*}{$24\times8$}} & \multicolumn{1}{c}{\multirow{3}[4]{*}{5}} \bigstrut\\
\multicolumn{1}{c|}{} & \multicolumn{1}{c|}{} & \multicolumn{1}{c|}{} & \multicolumn{1}{c}{} \\
\cline{2-3}
\multicolumn{1}{c|}{} & Max Pooling & $1\times1$ & \bigstrut\\
\bottomrule[1pt]
\end{tabular}
\end{center}
\label{tab:head}
\end{table}

\section{Experiment}
\subsection{Experimental Setting.}
\noindent\textbf{Datasets.} We conduct experiments on the CUHK-PEDES~\cite{li2017person} dataset, which is currently the only benchmark for person search by natural language. It contains 40,206 images of 13,003 different persons, where each image comes with two human-annotated sentences. The dataset is split into 11,003 identities with 34,054 images in the training set, 1,000 identities with 3,078 images in validation, and 1,000 identities with 3,074 images in testing set.

\noindent\textbf{Evaluation protocols.} Following the standard evaluation setting, we adopt Recall@K (K=1, 5, 10) as the retrieval criteria. Specifically, given a text description as query, Recall@K (R@K) reports the percentage of the images where at least one corresponding person is retrieved correctly among the top-K results.

\noindent\textbf{Implementation details.} For the global and local branches in image stream, we use the Basicblock as described in~\cite{he2016deep}, where each branch is randomly initialized (detailed architecture is shown in Table~\ref{tab:head}). We use horizontally flipping as data augmenting and resize all the images to $384\times128$.
We use the Adam solver as the training optimizer with weight decay set as $4\times10^{-5}$, and involves $64$ image-language pairs per mini-batch. The learning rate is initialized at $2\times10^{-4}$ for the first $40$ epochs during training, then decayed by a factor of $0.1$ for the remaining $30$ epochs. The whole experiment is implemented on a single Tesla V100 GPU machine. The hyperparameters in Eq.~\ref{eq:loss} are empirically set as: $\alpha=0.6,\beta=0.4,\tau_p=10,\tau_n=40$.

\noindent\textbf{Pedestrian attributes parsing.}
Based on the analysis of image and natural language annotations in the dataset, we warp both visual and textual attributes into $5$ categories: head (including descriptions related to hat, glasses, and face), clothes on the upper body, clothes on the lower body, shoes and bags (including backpack and handbag).
We reckon that these attributes are visually distinguishable from both modalities. In Figure~\ref{fig:annotation}, we visualize the segmentation maps generated by our human parsing network, where attribute regions can be properly segmented and associated with correct labels.

\subsection{Comparisons with the State-of-The-Arts}
\noindent\textbf{Result on CUHK-PEDES dataset.}
We summarize the performance of ViTAA and compare it with state-of-the-art methods in Table~\ref{tab:bylanguage} on the CUHK-PEDES test set. 
Methods like GNA-RNN~\cite{li2017person}, CMCE~\cite{li2017identity}, PWM-ATH~\cite{8354312} employ attention mechanism to learn the relation between visual and textual representation, while Dual Path~\cite{zheng2017dual}, CMPM+CMPC~\cite{zhang2018deep} design objective function for better joint embedding learning. These methods only learn and utilize the ``\textit{global}'' feature representation of both image and text. 
Moreover, MIA~\cite{niu2019improving} exploits ``\textit{region}'' information by dividing the input image into several horizontal stripes and extracting noun phrases from the natural language description. 
Similarly, GALM~\cite{jing2018pose} leverage ``\textit{keypoint}'' information from human pose estimation as an attention mechanism to assist feature learning and together with a noun phrases extractor implemented on input text.
Though the above two utilize the local-level representations, neither of them learns the associations between visual features with textual phrases. 
From Table~\ref{tab:bylanguage}, we observe that ViTAA shows a consistent lead on all metrics (R@1-10), outperforming the GALM~\cite{jing2018pose} by a margin of 1.85\%, 0.39\%, 0.55\% and claims the new state-of-the-art results. We note that though the performance could be considered as incremental, the shown improvement on the R@1 performance is challenging. It  suggests that the alignment learning of ViTAA contributes to the retrieval task directly. We further report the ablation studies on the effect of different components, and exhibit the attribute retrieval results quantitatively and qualitatively.

\begin{table}[t]
\caption{Person search results on the CUHK-PEDES test set. Best results are in bold.}
\begin{center}
\setlength{\tabcolsep}{7.5pt}
\begin{tabular}{c|c|ccc}
\toprule[1pt]
Method &Feature &R@1 &R@5 &R@10 \\
\hline
GNA-RNN~\cite{li2017person} &global &19.05 &- &53.64 \\
CMCE~\cite{li2017identity} &global &25.94 &- &60.48 \\
PWM-ATH~\cite{8354312} &global &27.14 &49.45 &61.02 \\
Dual Path~\cite{zheng2017dual} &global &44.40 &66.26 &75.07 \\
CMPM+CMPC~\cite{zhang2018deep} &global &49.37 &- &79.27 \\
MIA~\cite{niu2019improving} &global+region &53.10 &75.00 &82.90 \\
GALM~\cite{jing2018pose} &global+keypoint &54.12 &75.45 &82.97 \\
\hline
ViTAA &global+attribute &\textbf{55.97} &\textbf{75.84} &\textbf{83.52} \\
\bottomrule[1pt]
\end{tabular}
\end{center}
\label{tab:bylanguage}
\end{table}

\subsection{Ablation Study}
We carry out comprehensive ablations to evaluate the contribution of different components and the training configurations.

\noindent\textbf{Comparisons over different component combinations.}
To compare the individual contribution of each component, we set the baseline model as the one trained with only ID loss. In Table~\ref{tab:component}, we report the improvement of the proposed components (segmentation, global-alignment, and attribute-alignment) on the basis of the baseline model.
From the table, we have the following observations and analyses: First, using segmentation loss only brings marginal improvement because the visual features are not aligned with their corresponding textual features.
Similarly, we observe the same trend when the training is combined with only attribute-alignment loss where the visual features are not properly segmented, thus can not be associated for retrieval. 
An incremental gain is obtained by combining these two components.
Next, compared with attribute-level, global-level alignment greatly improves the performance under all criteria, which demonstrates the efficiency of the visual-textual alignment schema.
The cause of the performance gap is that: the former is learning the attribute similarity across different person identities while the latter is concentrating on the uniqueness of each person.
At the end, by combining all the loss terms yields the best performance, validating that our global-alignment and attribute-alignment learning are complimentary with each other. 

{
\setlength{\tabcolsep}{0.6em}
\begin{table}[t]
\caption{The improvement of components added on baseline model. Glb-Align and Attr-Align represent global-level and attribute-level alignment respectively.}
\begin{center}
\setlength{\tabcolsep}{7.5pt}
\begin{tabular}{c|c|c|ccc}
\toprule[1pt]
\multicolumn{3}{c|}{Model Component} &\multirow{2}{*}{R@1} & \multirow{2}{*}{R@5} &\multirow{2}{*}{R@10} \\
\cline{1-3}
Segmentation &Attr-Align &Glb-Align  \\
\hline
 & & &29.68 &51.84 &61.57 \\
\checkmark & & &30.93 &52.71 &63.11 \\
 &\checkmark & &31.40 &54.09 &63.66 \\
\checkmark &\checkmark & &39.26 &61.22 &68.14 \\
 & &\checkmark &52.27 &73.33 &81.61 \\
\checkmark &\checkmark &\checkmark &\textbf{55.97} &\textbf{75.84} &\textbf{83.52}\\
\bottomrule[1pt]
\end{tabular}
\end{center}
\label{tab:component}
\end{table}
}

\begin{figure}[t]
\centering
\includegraphics[width=1.0\textwidth]{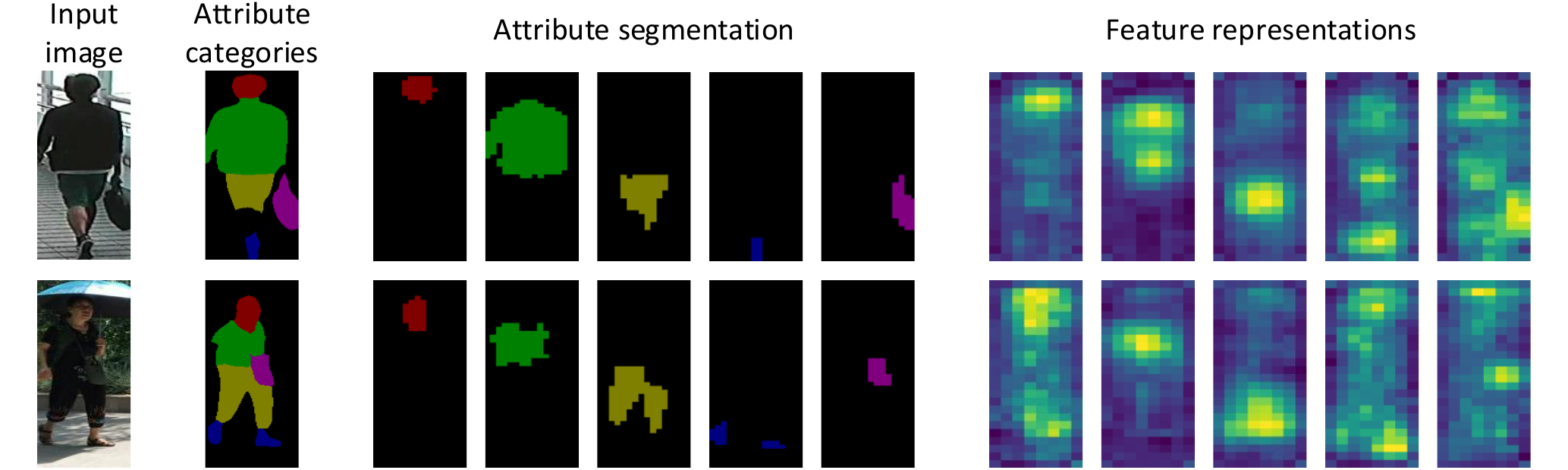}
\caption{From left to right, we exhibit the raw input person images, attribute labels generated by the pre-trained HRNet, attribute segmentation result from our segmentation layer, and their corresponded feature maps from the local branches.}
\label{fig:visana}
\end{figure}

\noindent\textbf{Visual attribute segmentation and representations.}
In Figure~\ref{fig:visana}, we visualize the segmentation maps from the segmentation layer and the feature representations of the local branches. It evidently shows that, even transferred using only a lightweight structure, the auxiliary person segmentation layer produces accurate pixel-wise labels under different human pose. This suggests that person parsing knowledge has been successfully distilled our local branches, which is crucial for the precise cross-modal alignment learning. On the right side of Figure~\ref{fig:visana}, we showcase the feature maps of local branch per attribute.

\begin{figure}[t]
\centering
\includegraphics[width=1.0\textwidth]{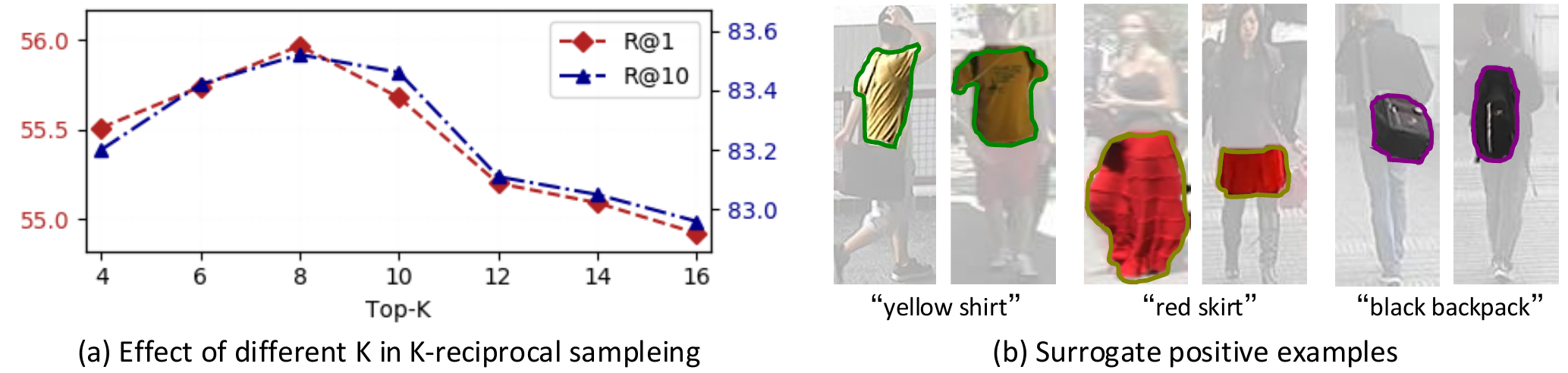}
\caption{(a) R@1 and R@10 results across different $K$ value in the proposed surrogate positive data sampling method. (b) Some examples of the surrogate positive data with different person identities.}
\label{fig:sample}
\end{figure}

\begin{figure}[t]
\centering
\includegraphics[width=1.0\textwidth]{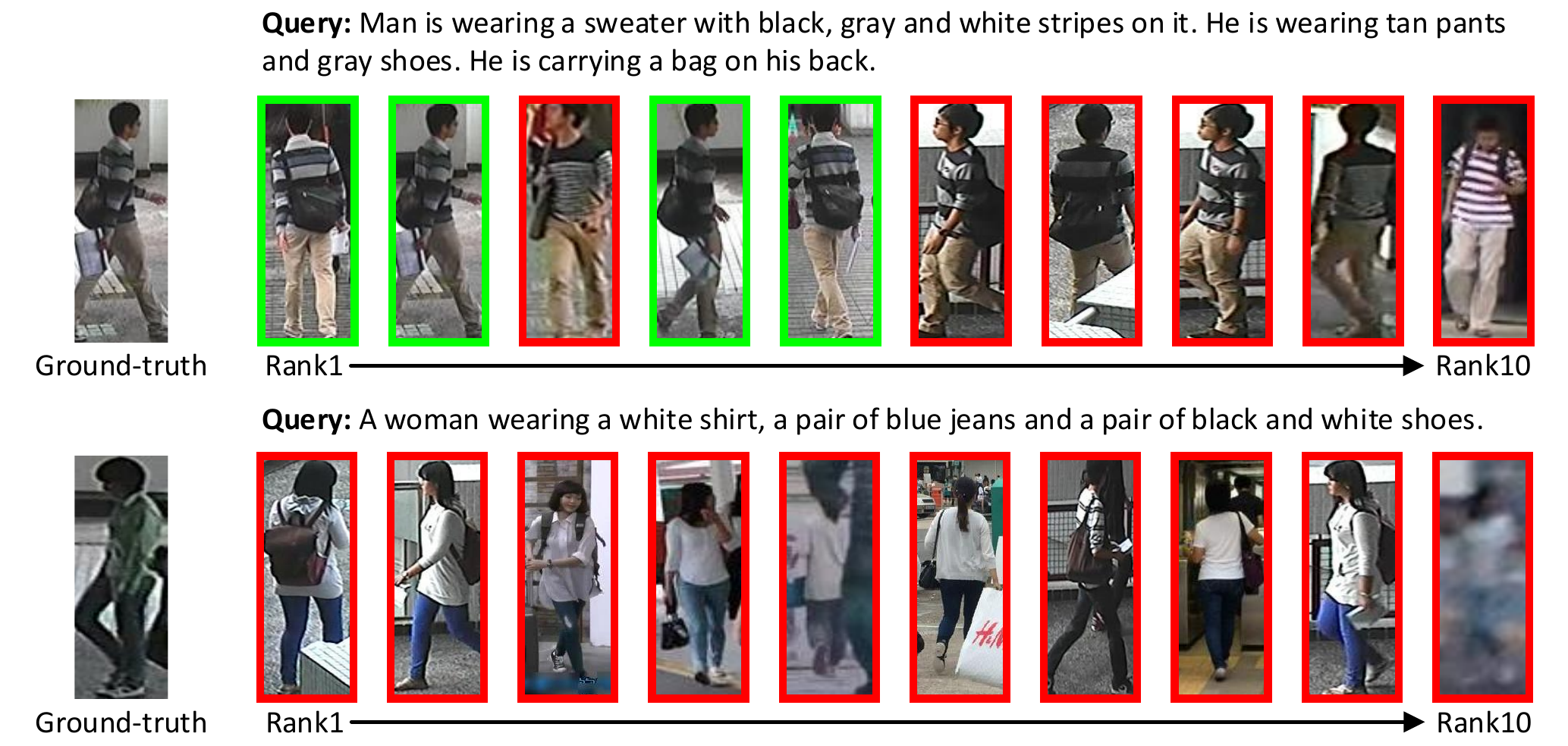}
\caption{Examples of person search results on CUHK-PEDES. We indicate the true/false matching results in \textcolor{green}{green}/\textcolor{red}{red} boxes.}
\label{fig:visana}
\end{figure}

\noindent\textbf{K-reciprocal sampling.}
We investigate how the value of $K$ impacts the pair-based sampling and learning process. We evaluate the R@1 and R@10 performance under different $K$ settings in Figure~\ref{fig:sample}(a). Ideally, the larger the $K$ is, the more potential surrogate positive samples will be mined, while this also comes with the possibility that more non-relevant examples (false positive examples) might be incorrectly sampled. Result in Figure~\ref{fig:sample}(a) agrees with our analysis: best R@1 and R@10 is achieved when $K$ is set to 8, and the performances are persistently declining as $K$ goes larger. In Figure~\ref{fig:sample}(b), we provide visual examinations of the surrogate positive pairs that mined by our sampling method. The visual attributes from different persons serve as valuable positive samples in our alignment learning schema.

\noindent\textbf{Qualitative analysis.}
We present the qualitative examples of person retrieval results to provide a more in-depth examination. As shown in Figure~\ref{fig:visana}, we illustrate the top-10 matching results using the given query. 
In the successful case (top), ViTAA precisely capture all attributes in the target person. It is worth noting that the wrong answers still capture the relevant attributes: ``sweater with black, gray and white stripes'', ``tan pants'', and ``carrying a bag''.
For the failure case (bottom), though the retrieved images are incorrect, we observe that all the attributes described in the query are there in almost all retrieved results.

\subsection{Extension: Attribute Retrieval}
In order to validate the ability of associating the visual attribute with the text phrase, we further conduct attribute retrieval experiment on the datasets of Market-1501~\cite{zheng2015scalable} and DukeMTMC~\cite{ristani2016performance}, where $27$ and $23$ human related attributes are annotated per image by~\cite{lin2019improving}. In our experiment, we use our pre-trained ViTAA on CUHK-PEDES without any further finetuning, and conduct the retrieval task using the attribute phrase as the query under R@1 and mAP metrics. In our experiment, we simply test on the \emph{upper-body clothing} attribute category, and post the retrieval results in Table~\ref{tab:attr_retrieve}. We introduce the details of our experiment in the supplementary materials. From Table~\ref{tab:attr_retrieve}, it clearly shows that ViTAA achieves great performances on almost all sub-attributes. This further strongly supports our argument that ViTAA is able to associate the visual attribute features with textual attribute descriptions successfully.

\begin{table}[t]
\caption{Upper-body clothing attribute retrieve results. Attr is the short form of attribute and ``upblack'' denotes the upper-body in black.}
\begin{center}
\setlength{\tabcolsep}{7.5pt}
{
\begin{tabular}{c|c|cc|c|c|cc}
\toprule[1pt]
\multicolumn{4}{c|}{Market1501} &\multicolumn{4}{c}{DukeMTMC} \\
\hline
Attr. &\#Target &R@1 &mAP &Attr. &\#Target &R@1 &mAP \\
\hline
upblack &1759 &99.2 &44.0 &upblack &11047 &100.0 &82.2 \\
upwhite &3491 &44.7 &64.8 &upwhite &1095 &91.3 &35.4 \\
upred &1354 &92.3 &54.8 &upred &821 &100.0 &44.6 \\
uppurple &363 &100.0 &61.2 &uppurple &65 &0.0 &9.0 \\
upyellow &1195 &72.7 &75.6 &upgray &2012 &81.7 &29.8 \\
upgray &1755 &49.7 &55.2 &upblue &1577 &77.9 &31.3 \\
upblue &1100 &70.1 &32.4 &upgreen &417 &89.2 &24.5 \\
upgreen &949 &87.9 &50.4 &upbrown &345 &18.3 &15.2 \\
\bottomrule[1pt]
\end{tabular}
}
\end{center}
\label{tab:attr_retrieve}
\end{table}
\section{Conclusion}
In this work, we present a novel ViTAA model to address the person search by natural language task from the perspective of an attribute-specific alignment learning. In contrast to the existing methods, ViTAA fully exploits the common attribute information in both visual and textual modalities across different person identities, and further builds strong association between the visual attribute features and their corresponding textual phrases by using our alignment learning schema. We show that ViTAA achieves state-of-the-art results on the challenging benchmark CUHK-PEDES and demonstrate its promising potential that further advances the person search by natural language domain.

\noindent\textbf{Acknowledgements.}
Vising scholarship support for Z. Wang from the China Scholarship Council \#201806020020 and Amazon AWS Machine Learning Research Award (MLRA) support are greatly appreciated. Any opinions, findings, and conclusion or recommendations expressed in this material are those of the authors and do not necessarily reflect the view of the sponsors.

%
%
\bibliographystyle{splncs04}
\bibliography{egbib}

\begin{thebibliography}{10}
\providecommand{\url}[1]{\texttt{#1}}
\providecommand{\urlprefix}{URL }
\providecommand{\doi}[1]{https://doi.org/#1}

\bibitem{antol2015vqa}
Antol, S., Agrawal, A., Lu, J., Mitchell, M., Batra, D., Lawrence~Zitnick, C.,
  Parikh, D.: Vqa: Visual question answering. In: Proceedings of the IEEE
  international conference on computer vision. pp. 2425--2433 (2015)

\bibitem{benenson2014ten}
Benenson, R., Omran, M., Hosang, J., Schiele, B.: Ten years of pedestrian
  detection, what have we learned? In: European Conference on Computer Vision.
  pp. 613--627. Springer (2014)

\bibitem{chen2018improving}
Chen, D., Li, H., Liu, X., Shen, Y., Shao, J., Yuan, Z., Wang, X.: Improving
  deep visual representation for person re-identification by global and local
  image-language association. In: Proceedings of the European Conference on
  Computer Vision (ECCV). pp. 54--70 (2018)

\bibitem{8354312}
{Chen}, T., {Xu}, C., {Luo}, J.: Improving text-based person search by spatial
  matching and adaptive threshold. In: 2018 IEEE Winter Conference on
  Applications of Computer Vision (WACV). pp. 1879--1887 (March 2018)

\bibitem{dollar2009pedestrian}
Doll{\'a}r, P., Wojek, C., Schiele, B., Perona, P.: Pedestrian detection: A
  benchmark. In: Computer Vision and Pattern Recognition, 2009. CVPR 2009. IEEE
  Conference on. pp. 304--311. IEEE (2009)

\bibitem{Dong_2019_ICCV}
Dong, Q., Gong, S., Zhu, X.: Person search by text attribute query as zero-shot
  learning. In: The IEEE International Conference on Computer Vision (ICCV)
  (October 2019)

\bibitem{fang2020video2commonsense}
Fang, Z., Gokhale, T., Banerjee, P., Baral, C., Yang, Y.: Video2commonsense:
  Generating commonsense descriptions to enrich video captioning. arXiv
  preprint arXiv:2003.05162  (2020)

\bibitem{Fang_2019_CVPR}
Fang, Z., Kong, S., Fowlkes, C., Yang, Y.: Modularized textual grounding for
  counterfactual resilience. In: The IEEE Conference on Computer Vision and
  Pattern Recognition (CVPR) (June 2019)

\bibitem{fang2018weakly}
Fang, Z., Kong, S., Yu, T., Yang, Y.: Weakly supervised attention learning for
  textual phrases grounding. arXiv preprint arXiv:1805.00545  (2018)

\bibitem{frome2013devise}
Frome, A., Corrado, G.S., Shlens, J., Bengio, S., Dean, J., Ranzato, M.,
  Mikolov, T.: Devise: A deep visual-semantic embedding model. In: Advances in
  neural information processing systems. pp. 2121--2129 (2013)

\bibitem{garcia2015person}
Garcia, J., Martinel, N., Micheloni, C., Gardel, A.: Person re-identification
  ranking optimisation by discriminant context information analysis. In:
  Proceedings of the IEEE International Conference on Computer Vision. pp.
  1305--1313 (2015)

\bibitem{goldberg2014word2vec}
Goldberg, Y., Levy, O.: word2vec explained: deriving mikolov et al.'s
  negative-sampling word-embedding method. arXiv preprint arXiv:1402.3722
  (2014)

\bibitem{Gong:2014:PR:2584512}
Gong, S., Cristani, M., Yan, S., Loy, C.C.: Person Re-Identification. Springer
  Publishing Company, Incorporated (2014)

\bibitem{Guo_2019_ICCV}
Guo, J., Yuan, Y., Huang, L., Zhang, C., Yao, J.G., Han, K.: Beyond human
  parts: Dual part-aligned representations for person re-identification. In:
  The IEEE International Conference on Computer Vision (ICCV) (October 2019)

\bibitem{han2019re}
Han, C., Ye, J., Zhong, Y., Tan, X., Zhang, C., Gao, C., Sang, N.: Re-id driven
  localization refinement for person search. In: Proceedings of the IEEE
  International Conference on Computer Vision. pp. 9814--9823 (2019)

\bibitem{he2017mask}
He, K., Gkioxari, G., Doll{\'a}r, P., Girshick, R.: Mask r-cnn. In: Proceedings
  of the IEEE international conference on computer vision. pp. 2961--2969
  (2017)

\bibitem{he2016deep}
He, K., Zhang, X., Ren, S., Sun, J.: Deep residual learning for image
  recognition. In: Proceedings of the IEEE conference on computer vision and
  pattern recognition. pp. 770--778 (2016)

\bibitem{jeon2003automatic}
Jeon, J., Lavrenko, V., Manmatha, R.: Automatic image annotation and retrieval
  using cross-media relevance models. In: Proceedings of the 26th annual
  international ACM SIGIR conference on Research and development in informaion
  retrieval. pp. 119--126 (2003)

\bibitem{jing2018pose}
Jing, Y., Si, C., Wang, J., Wang, W., Wang, L., Tan, T.: Pose-guided joint
  global and attentive local matching network for text-based person search.
  arXiv preprint arXiv:1809.08440  (2018)

\bibitem{kalayeh2018human}
Kalayeh, M.M., Basaran, E., G{\"o}kmen, M., Kamasak, M.E., Shah, M.: Human
  semantic parsing for person re-identification. In: Proceedings of the IEEE
  Conference on Computer Vision and Pattern Recognition. pp. 1062--1071 (2018)

\bibitem{karpathy2015deep}
Karpathy, A., Fei-Fei, L.: Deep visual-semantic alignments for generating image
  descriptions. In: Proceedings of the IEEE conference on computer vision and
  pattern recognition. pp. 3128--3137 (2015)

\bibitem{klein2003fast}
Klein, D., Manning, C.D.: Fast exact inference with a factored model for
  natural language parsing. In: Advances in neural information processing
  systems. pp. 3--10 (2003)

\bibitem{layne2014attributes}
Layne, R., Hospedales, T.M., Gong, S.: Attributes-based re-identification. In:
  Person Re-Identification, pp. 93--117. Springer (2014)

\bibitem{li2017identity}
Li, S., Xiao, T., Li, H., Yang, W., Wang, X.: Identity-aware textual-visual
  matching with latent co-attention. In: Proceedings of the IEEE International
  Conference on Computer Vision. pp. 1890--1899 (2017)

\bibitem{li2017person}
Li, S., Xiao, T., Li, H., Zhou, B., Yue, D., Wang, X.: Person search with
  natural language description. In: Proceedings of the IEEE Conference on
  Computer Vision and Pattern Recognition. pp. 1970--1979 (2017)

\bibitem{li2014deepreid}
Li, W., Zhao, R., Xiao, T., Wang, X.: Deepreid: Deep filter pairing neural
  network for person re-identification. In: Proceedings of the IEEE conference
  on computer vision and pattern recognition. pp. 152--159 (2014)

\bibitem{liang2018look}
Liang, X., Gong, K., Shen, X., Lin, L.: Look into person: Joint body parsing \&
  pose estimation network and a new benchmark. IEEE transactions on pattern
  analysis and machine intelligence  \textbf{41}(4),  871--885 (2018)

\bibitem{ATR}
Liang, X., Liu, S., Shen, X., Yang, J., Liu, L., Dong, J., Lin, L., Yan, S.:
  Deep human parsing with active template regression. Pattern Analysis and
  Machine Intelligence, IEEE Transactions on (12),  2402--2414 (Dec 2015)

\bibitem{lin2019improving}
Lin, Y., Zheng, L., Zheng, Z., Wu, Y., Hu, Z., Yan, C., Yang, Y.: Improving
  person re-identification by attribute and identity learning. Pattern
  Recognition  (2019)

\bibitem{liu2017hydraplus}
Liu, X., Zhao, H., Tian, M., Sheng, L., Shao, J., Yi, S., Yan, J., Wang, X.:
  Hydraplus-net: Attentive deep features for pedestrian analysis. In:
  Proceedings of the IEEE international conference on computer vision. pp.
  350--359 (2017)

\bibitem{manning-EtAl:2014:P14-5}
Manning, C.D., Surdeanu, M., Bauer, J., Finkel, J., Bethard, S.J., McClosky,
  D.: The {Stanford} {CoreNLP} natural language processing toolkit. In:
  Association for Computational Linguistics (ACL) System Demonstrations. pp.
  55--60 (2014)

\bibitem{niu2019improving}
Niu, K., Huang, Y., Ouyang, W., Wang, L.: Improving description-based person
  re-identification by multi-granularity image-text alignments. arXiv preprint
  arXiv:1906.09610  (2019)

\bibitem{plummer2015flickr30k}
Plummer, B.A., Wang, L., Cervantes, C.M., Caicedo, J.C., Hockenmaier, J.,
  Lazebnik, S.: Flickr30k entities: Collecting region-to-phrase correspondences
  for richer image-to-sentence models. In: Proceedings of the IEEE
  international conference on computer vision. pp. 2641--2649 (2015)

\bibitem{ristani2016performance}
Ristani, E., Solera, F., Zou, R., Cucchiara, R., Tomasi, C.: Performance
  measures and a data set for multi-target, multi-camera tracking. In: European
  Conference on Computer Vision. pp. 17--35. Springer (2016)

\bibitem{rohrbach2016grounding}
Rohrbach, A., Rohrbach, M., Hu, R., Darrell, T., Schiele, B.: Grounding of
  textual phrases in images by reconstruction. In: European Conference on
  Computer Vision. pp. 817--834. Springer (2016)

\bibitem{shekhar2012word}
Shekhar, R., Jawahar, C.: Word image retrieval using bag of visual words. In:
  2012 10th IAPR International Workshop on Document Analysis Systems. pp.
  297--301. IEEE (2012)

\bibitem{si2018dual}
Si, J., Zhang, H., Li, C.G., Kuen, J., Kong, X., Kot, A.C., Wang, G.: Dual
  attention matching network for context-aware feature sequence based person
  re-identification. In: Proceedings of the IEEE Conference on Computer Vision
  and Pattern Recognition. pp. 5363--5372 (2018)

\bibitem{su2017pose}
Su, C., Li, J., Zhang, S., Xing, J., Gao, W., Tian, Q.: Pose-driven deep
  convolutional model for person re-identification. In: Proceedings of the IEEE
  International Conference on Computer Vision. pp. 3960--3969 (2017)

\bibitem{su2018multi}
Su, C., Zhang, S., Xing, J., Gao, W., Tian, Q.: Multi-type attributes driven
  multi-camera person re-identification. Pattern Recognition  \textbf{75},
  77--89 (2018)

\bibitem{sudowe2015person}
Sudowe, P., Spitzer, H., Leibe, B.: Person attribute recognition with a
  jointly-trained holistic cnn model. In: Proceedings of the IEEE International
  Conference on Computer Vision Workshops. pp. 87--95 (2015)

\bibitem{suh2018part}
Suh, Y., Wang, J., Tang, S., Mei, T., Mu~Lee, K.: Part-aligned bilinear
  representations for person re-identification. In: Proceedings of the European
  Conference on Computer Vision (ECCV). pp. 402--419 (2018)

\bibitem{Sun_2019_CVPR}
Sun, K., Xiao, B., Liu, D., Wang, J.: Deep high-resolution representation
  learning for human pose estimation. In: The IEEE Conference on Computer
  Vision and Pattern Recognition (CVPR) (June 2019)

\bibitem{sun2018beyond}
Sun, Y., Zheng, L., Yang, Y., Tian, Q., Wang, S.: Beyond part models: Person
  retrieval with refined part pooling (and a strong convolutional baseline).
  In: Proceedings of the European Conference on Computer Vision (ECCV). pp.
  480--496 (2018)

\bibitem{tan2019attention}
Tan, Z., Yang, Y., Wan, J., Hang, H., Guo, G., Li, S.Z.: Attention-based
  pedestrian attribute analysis. IEEE transactions on image processing (12),
  6126--6140 (2019)

\bibitem{wang2018mancs}
Wang, C., Zhang, Q., Huang, C., Liu, W., Wang, X.: Mancs: A multi-task
  attentional network with curriculum sampling for person re-identification.
  In: Proceedings of the European Conference on Computer Vision (ECCV). pp.
  365--381 (2018)

\bibitem{wang2018learning}
Wang, G., Yuan, Y., Chen, X., Li, J., Zhou, X.: Learning discriminative
  features with multiple granularities for person re-identification. In: 2018
  ACM Multimedia Conference on Multimedia Conference. pp. 274--282. ACM (2018)

\bibitem{wang2020resisting}
Wang, Z., Wang, J., Yang, Y.: Resisting crowd occlusion and hard negatives for
  pedestrian detection in the wild. arXiv preprint arXiv:2005.07344  (2020)

\bibitem{wen2016discriminative}
Wen, Y., Zhang, K., Li, Z., Qiao, Y.: A discriminative feature learning
  approach for deep face recognition. In: European conference on computer
  vision. pp. 499--515. Springer (2016)

\bibitem{Wu_2019_CVPR}
Wu, H., Mao, J., Zhang, Y., Jiang, Y., Li, L., Sun, W., Ma, W.Y.: Unified
  visual-semantic embeddings: Bridging vision and language with structured
  meaning representations. In: The IEEE Conference on Computer Vision and
  Pattern Recognition (CVPR) (June 2019)

\bibitem{xu2018attention}
Xu, J., Zhao, R., Zhu, F., Wang, H., Ouyang, W.: Attention-aware compositional
  network for person re-identification. In: Proceedings of the IEEE Conference
  on Computer Vision and Pattern Recognition. pp. 2119--2128 (2018)

\bibitem{xu2015show}
Xu, K., Ba, J., Kiros, R., Cho, K., Courville, A., Salakhudinov, R., Zemel, R.,
  Bengio, Y.: Show, attend and tell: Neural image caption generation with
  visual attention. In: International conference on machine learning. pp.
  2048--2057 (2015)

\bibitem{ijcai2018-153}
Yin, Z., Zheng, W.S., Wu, A., Yu, H.X., Wan, H., Guo, X., Huang, F., Lai, J.:
  Adversarial attribute-image person re-identification. In: Proceedings of the
  Twenty-Seventh International Joint Conference on Artificial Intelligence,
  {IJCAI-18}. pp. 1100--1106. International Joint Conferences on Artificial
  Intelligence Organization (7 2018)

\bibitem{you2018end}
You, Q., Zhang, Z., Luo, J.: End-to-end convolutional semantic embeddings. In:
  Proceedings of the IEEE Conference on Computer Vision and Pattern
  Recognition. pp. 5735--5744 (2018)

\bibitem{zhang2017range}
Zhang, X., Fang, Z., Wen, Y., Li, Z., Qiao, Y.: Range loss for deep face
  recognition with long-tailed training data. In: Proceedings of the IEEE
  International Conference on Computer Vision. pp. 5409--5418 (2017)

\bibitem{zhang2018deep}
Zhang, Y., Lu, H.: Deep cross-modal projection learning for image-text
  matching. In: Proceedings of the European Conference on Computer Vision
  (ECCV). pp. 686--701 (2018)

\bibitem{zhang2019densely}
Zhang, Z., Lan, C., Zeng, W., Chen, Z.: Densely semantically aligned person
  re-identification. In: Proceedings of the IEEE Conference on Computer Vision
  and Pattern Recognition. pp. 667--676 (2019)

\bibitem{zhao2018understanding}
Zhao, J., Li, J., Cheng, Y., Sim, T., Yan, S., Feng, J.: Understanding humans
  in crowded scenes: Deep nested adversarial learning and a new benchmark for
  multi-human parsing. In: 2018 ACM Multimedia Conference on Multimedia
  Conference. pp. 792--800. ACM (2018)

\bibitem{zheng2019pose}
Zheng, L., Huang, Y., Lu, H., Yang, Y.: Pose invariant embedding for deep
  person re-identification. IEEE Transactions on Image Processing  (2019)

\bibitem{zheng2015scalable}
Zheng, L., Shen, L., Tian, L., Wang, S., Wang, J., Tian, Q.: Scalable person
  re-identification: A benchmark. In: Proceedings of the IEEE international
  conference on computer vision. pp. 1116--1124 (2015)

\bibitem{zheng2017dual}
Zheng, Z., Zheng, L., Garrett, M., Yang, Y., Shen, Y.D.: Dual-path
  convolutional image-text embedding with instance loss. arXiv preprint
  arXiv:1711.05535  (2017)

\bibitem{zhong2017re}
Zhong, Z., Zheng, L., Cao, D., Li, S.: Re-ranking person re-identification with
  k-reciprocal encoding. In: Proceedings of the IEEE Conference on Computer
  Vision and Pattern Recognition. pp. 1318--1327 (2017)

\end{thebibliography}
\end{document}